# Automated Parking Trajectory Generation Using Deep Reinforcement Learning

Zheyu Zhang
Independent Researcher
Beijing, China
zheyuz2980@gmail.com

Yutong Luo
Independent Researcher
Beijing, China
yutongluo0110@gmail.com

Yongzhou Chen
Independent Researcher
Beijing, China
yongzhouc@outlook.com

Haopeng Zhao
Independent Researcher
Beijing, China
haopeng.zhao1894@gmail.com

Zhichao Ma
Independent Researcher
Shanghai, China
ma.zhi.chao.max@gmail.com

Hao Liu*
Independent Researcher
Beijing, China
modiy.lu@gmail.com

*Abstract*—**Autonomous parking is a key technology in modern autonomous driving systems, requiring high precision, strong adaptability, and efficiency in complex environments. This paper proposes a Deep Reinforcement Learning (DRL) framework based on the Soft Actor-Critic (SAC) algorithm to optimize autonomous parking tasks. SAC, an off-policy method with entropy regularization, is particularly well-suited for continuous action spaces, enabling fine-grained vehicle control. We model the parking task as a Markov Decision Process (MDP) and train an agent to maximize cumulative rewards while balancing exploration and exploitation through entropy maximization. The proposed system integrates multiple sensor inputs into a high-dimensional state space and leverages SAC's dual critic networks and policy network to achieve stable learning. Simulation results show that the SAC-based approach delivers high parking success rates, reduced maneuver times, and robust handling of dynamic obstacles, outperforming traditional rule-based methods and other DRL algorithms. This study demonstrates SAC's potential in autonomous parking and lays the foundation for real-world applications.**

## I. INTRODUCTION

With the rapid and breakthrough development of autonomous driving technology, automatic parking has become an important research focus, aiming to significantly enhance the automation capabilities of intelligent vehicles[1][2]. As cities become more crowded and parking spaces become scarce, there is an increasing demand for vehicles that can independently navigate and perform parking tasks in a variety of demanding situations[3]. Traditional parking algorithms have long been the cornerstone of automatic parking systems, typically relying on techniques such as geometric path planning, fuzzy control, or proportional-integral-derivative (PID) controllers[4]. While these traditional methods work in simple or predictable environments, they often fail when faced with the fluidity of dynamic environments or the subtle challenges of complex parking scenarios[5][6]. Their reliance on rigid preset rules and assumptions hinders their flexibility, highlighting the urgent need for more sophisticated, adaptable, and intelligent solutions to cope with the complexity of real-world driving[7].

In recent years, the rise of deep reinforcement learning (DRL) has opened up an exciting and efficient path to address these shortcomings, providing a compelling solution for automatic parking through its ability to promote adaptive and self-evolving control strategies. Reference [8] showed that combining different deep learning models with an ensemble stacking approach can greatly improve model prediction accuracy. This work can harness the strengths of multiple deep learning techniques to better capture patterns in data. Reference [9] demonstrated the effectiveness of combining multiple machine learning models, including Neural Networks, LightGBM, and Random Forest, into an ensemble framework for credit score prediction. This work can significantly enhance model performance on imbalanced datasets by improving precision, recall, and overall robustness.

By combining the strengths of deep neural networks with the principles of reinforcement learning, DRL allows a system to improve its behavior through iterative experience, learning to smartly respond to new and unforeseen situations without tedious manual coding[10]. This paper leverages a cutting-edge automated parking approach based on the soft actor-critic (SAC) algorithm, a prominent approach in the DRL family, and rigorously tests its practicality and performance through detailed simulation experiments[11][12]. These simulations are designed to reflect real-life parking challenges, providing a powerful platform to evaluate how SAC-based systems handle the unpredictable and multifaceted nature of modern parking tasks[13][14][15].

Traditional parking methods include a range of established techniques, each with their own strengths and limitations[16][17]. Path planning-based methods, such as the A* algorithm and the rapidly exploring random trees (RRT) algorithm, excel at designing feasible routes by treating parking as a geometric puzzle, plotting precise trajectories from a starting location to a specified location while accounting for vehicle dynamics and spatial constraints. However, these methods rely on accurate, unchanging maps of the surrounding environment, and their effectiveness decreases when conditions change unexpectedly, such as in busy urban environments. Thanks to its entropy regularization framework, the SAC algorithm excels particularly at continuous control tasks, such as the smooth, precise adjustments required for steering and speed during parking. This framework not only promotes exploration, but also ensures policy stability, preventing it from locking into suboptimal modes too early. In this study, we leverage the strengths of SAC to enhance automated parking operations, aiming to create a system that can learn, adapt, and reliably perform across a variety of parking requirements, and validated through extensive simulations that push its limits in realistic, dynamic environments.

## II. METHODOLOGY

The challenge of autonomous parking requires a framework capable of translating high-dimensional sensory inputs into precise, continuous control actions under varying environmental conditions. To address this, we conceptualize the parking task as a Markov Decision Process (MDP), which provides a structured way to model the sequential decision-making inherent in maneuvering a vehicle into a designated spot. The MDP is characterized by a state space $S$ that encapsulates critical vehicle and environmental information, such as the vehicle's position coordinates $(x, y)$, its orientation $\theta$, its velocity $v$, and a rich set of sensor data including LiDAR distance measurements and camera-derived features. The action space $A$ is continuous, comprising the steering angle $\delta$, constrained within $[-\delta_{\max}, \delta_{\max}]$, and the throttle/brake input $u$, ranging from $[-1,1]$ to represent full braking to full acceleration. The reward function $R(s,a,s')$ is crafted to guide the learning process, rewarding the agent for reducing the distance to the target parking location while penalizing collisions, orientation misalignment, and excessive time consumption. The transition dynamics $P(s'|s,a)$ are not explicitly modeled but are implicitly captured through interactions within a simulated environment, and a discount factor $\gamma \in [0,1)$ balances immediate versus long-term rewards. The ultimate objective is to derive an optimal policy $pi(a|s)$ that maximizes the expected cumulative reward, enabling the vehicle to park efficiently and accurately.

To achieve this, we employ the Soft Actor-Critic (SAC) algorithm, a state-of-the-art deep reinforcement learning method particularly suited for continuous control tasks like autonomous parking. Fig. 1 shows the overall structure. Unlike traditional RL approaches that focus solely on reward maximization, SAC introduces an entropy-regularized objective that encourages exploration by maximizing both the expected reward and the entropy of the policy. This is formally expressed as

$$J(\pi) = \mathbb{E}_{\tau \sim \pi}\left[\sum_{t=0}^{\infty} \gamma^t \left(R(s_t, a_t, s_{t+1}) + \alpha \mathcal{H}(\pi(\cdot | s_t))\right)\right]$$

where $\mathcal{H}(\pi(\cdot|s_t))$ represents the policiy's entropy at state $s_t$, and $\alpha$ is a temperature parameter that dynamically adjusts the trade-off between exploration and exploitation. The SAC framework relies on a trio of neural networks working in tandem: a policy network and two Q-value networks. The policy network, parameterized by $\phi$ and denoted $\pi_\phi(a|s)$, outputs a Gaussian distribution over actions, with the mean and variance learned to represent the steering angle and throttle input. This network is a multi-layer perceptron (MLP) with three hidden layers of 256 units and ReLU activations, taking as input the concatenated state and action vectors and outputting a scalar $Q$-value. To stabilize training, we maintain target Q-networks $Q_{\theta_1}(s,a)$ and $Q_{\theta_2}(s,a)$, parameterized by $\theta_1$ and $\theta_2$, estimate the soft $Q$-values and are designed to combat overestimation bias by taking the minimum of their predictions during policy updates. Each Q-network is also an MLP with three hidden layers of 256 units and ReLU activations, taking as input the concatenated state and action vectors and outputting a scalar $Q$-value. To stabilize training, we maintain target Q-networks $Q_{\bar{\theta}1}$ and $Q_{\bar{\theta}_2}$, which are slowly updated via Polyak averaging. The Q-networks are trained by minimizing the soft Bellman residual,

$$J_Q(\theta_i) = \mathbb{E}_{(s,a,s',r) \sim \mathcal{D}}\left[\left(Q_{\theta_i}(s,a) - \left(r + \gamma\left(\min_{j=1,2} Q_{\bar{\theta}_j}(s',a') - \alpha \log \pi_\phi(a'|s')\right)\right)\right)^2\right]$$

where $\mathcal{D}$ is a replay buffer storing past transitions, and $a' \sim \pi_\phi(\cdot|s')$ is sampled from the current policy. The policy network, in turn, is optimized to minimize the KL-divergence between itself and the exponential of the $Q$-function, approximated as

$$J_\pi(\phi) = \mathbb{E}_{s \sim \mathcal{D}}\left[\mathbb{E}_{a \sim \pi_\phi}\left[\alpha \log \pi_\phi(a|s) - Q_{\theta_{\min}}(s,a)\right]\right]$$

with $\alpha$ adaptively tuned to maintain a target entropy level, enhancing robustness in the face of environmental uncertainty.

The reward function is a critical component of the learning process, designed to balance multiple objectives in parking. We define it as

$$R(s,a,s') = w_1 \cdot (-\text{dist}(s', s_{\text{goal}})) + w_2 \cdot (-\Delta\theta) \\ - w_3 \cdot \text{collision} - w_4 \cdot \text{time}$$

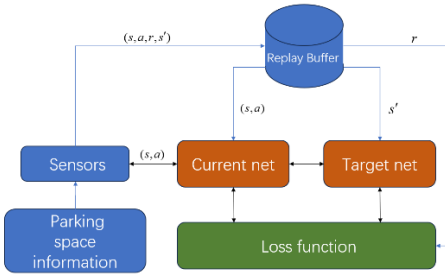

Fig 1. Overall structure.

where $\text{dist}(s', s_{\text{goal}})$ measures the Euclidean distance from the vehicle's new state $s'$ to the target parking spot, $\Delta\theta$ quantifies the orientation error relative to the desired alignment, collision is a binary penalty, and time penalizes prolonged maneuvers. The weights $w_1$, $w_2$, $w_3$, $w_4$ are empirically tuned to prioritize parking accuracy while discouraging unsafe or inefficient behavior. This Formulation ensures that the agent learns to position the vehicle close to the target and align it properly without colliding with obstacles or taking excessive steps. To train the SAC agent, we developed a custom simulation environment based on a kinematic bicycle model, which approximates vehicle dynamics as

$$x_{t+1} = x_t + v_t \cos(\theta_t)\Delta t$$
$$y_{t+1} = y_t + v_t \sin(\theta_t)\Delta t$$
$$\theta_{t+1} = \theta_t + \frac{v_t}{L}\tan(\delta_t)\Delta t$$

where $L$ is the wheelbase, and $\Delta t$ is the timestep. The environment simulates a $20\text{m} \times 20\text{m}$ parking lot with a $4\text{m} \times 2\text{m}$ target spot, populated with static obstacles and occasional dynamic objects. Sensor inputs are simulated with noise—LiDAR rays perturbed by Gaussian noise and camera data represented as low-resolution occupancy grids—mimicking real-world imperfections. This setup allows the SAC agent to learn robust parking strategies by interacting with a realistic yet controllable proxy for physical systems, leveraging the replay buffer to sample diverse experiences and refine its policy and $Q$-functions iteratively.

The structure of a trajectory planning system leveraging deep reinforcement learning is depicted in the Fig 1. Upon obtaining details about the parking area and vehicle specifications, the reinforcement learning algorithm evaluates the current environmental state to make informed decisions, producing a path from the initial position to the destination. The state is subsequently refined using a vehicle dynamics model. In contrast to conventional approaches, such as optimization-driven online planning, which demands iterative nonlinear computations, or sampling-based techniques that rely heavily on continuous collision checks, these methods often incur significant computational delays. By harnessing the capabilities of a trained reinforcement learning model, a preliminary trajectory from start to finish can be generated swiftly. This approach capitalizes on the model's learned patterns, enabling rapid and adaptable path planning in intricate scenarios, thereby offering a more efficient alternative for real-time applications.

## III. EXPERIMENTS

To validate the effectiveness of the SAC algorithm for autonomous parking, we conducted simulation experiments using Gazebo 11.0 integrated with ROS Noetic to simulate realistic vehicle dynamics and sensor data, approximating real-world conditions. The test vehicle controlled via continuous steering angle (-30° to 30°) and throttle/brake (-1 to 1), aligning with SAC's continuous action space. Control commands and sensor data were exchanged through ROS 1 topics for real-time interaction.

Three scenarios were designed: (1) parallel parking; (2) perpendicular parking; and (3) complex mixed parking. Each scenario varied initial vehicle positions and orientations to increase complexity.

The SAC algorithm, implemented in PyTorch, used a policy network and value network, each with three fully connected layers. The policy network outputted action means and standard deviations for steering and throttle, while the value network estimated Q-values. The state space included vehicle position (x, y), orientation (θ), speed (v), distance to the parking space, obstacle distances, and parking space geometry. The action space was a two-dimensional vector for steering and throttle/brake. Table I shows the training parameters of SAC.

TABLE I. PARAMETER OF SAC

| Parameter | Description | Value |
| --- | --- | --- |
| $M$ | Total episode | 3000 |
| $T$ | Timesteps | 1000 |
| $B$ | Batch size | 128 |
| $\gamma$ | Reward discount | 0.99 |
| $\sigma$ | Noise variance | 0.01 |
| $\tau$ | Smoothing coefficient | 0.05 |

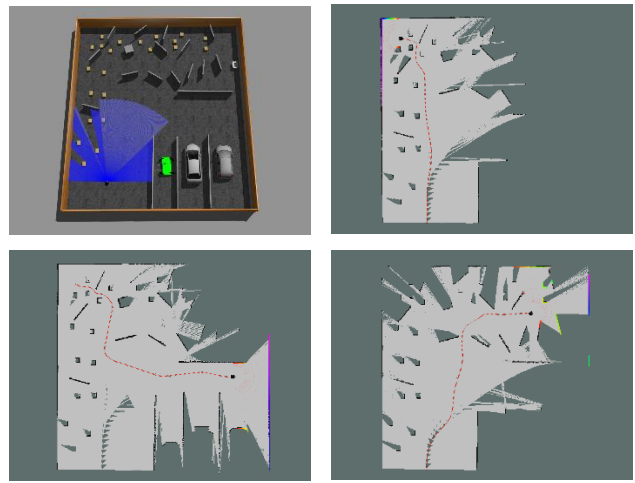

Fig 2. Experimental environment and cases.

Fig. 2 illustrates the experimental setup constructed in the Gazebo physical simulation environment, along with the outcomes of three distinct path planning simulations. To evaluate the performance of path planning algorithms, we designed a complex, randomly generated environment featuring various obstacles and terrain characteristics to closely mimic real-world navigation challenges. The experiments involved conducting multiple path planning tasks with randomly selected start and end points, aimed at assessing the algorithms' robustness and efficiency under diverse initial conditions.

The simulation environment was developed in Gazebo, leveraging its physics engine to accurately model the interactions between the robot and the environment, including collision detection, friction, and gravity. The obstacles within the scene were randomly generated in terms of position, size, and shape to ensure the generalizability of the experiments and prevent algorithmic overfitting to specific scenarios. Similarly, the start and end points were randomly distributed within navigable areas to ensure the path planning tasks were sufficiently diverse and representative.

The three path planning results shown in Figure 1 correspond to outputs from different algorithms or parameter configurations. Each path successfully navigates from the start to the end point, avoiding all obstacles while optimizing metrics such as path length or safety. The variations among the paths highlight the algorithms' performance in handling random environments, particularly in scenarios involving narrow passages, complex terrains.

TABLE II. PLANNING TIME OF DIFFERENT ALGORITHMS

| Method | Case 1 | Case 2 | Case 3 |
|---|---|---|---|
| Hybrid A* | 15.46 s | 1.36 s | 4.61 s |
| DQN | 13.64 s | 3.31 s | 5.87 s |
| DDQN | 14.25 s | 2.84 s | 6.35 s |
| SAC | 11.23 s | 2.43 s | 4.27 s |

Table II presents a comparative analysis of the time performance of various path planning algorithms across different scenarios, specifically including Hybrid A*, DQN, DDQN, and SAC algorithms. By thoroughly comparing the performance of these algorithms in multiple test scenarios, we can clearly observe the differences in their path generation efficiency. The experimental results demonstrate that the SAC algorithm significantly outperforms other methods in terms of path generation speed, highlighting its efficiency in complex environments.

IV. CONCLUSIONS

Through extensive simulation experiments, the SAC-based approach demonstrated superior performance in navigating a variety of parking scenarios, including parallel, perpendicular, and complex mixed configurations, with notable success rates, precise trajectory control, and resilience to obstacles. The algorithm's entropy-regularized framework effectively facilitated a balance between exploration and exploitation, enabling the development of adaptive and self-learning control policies that surpassed traditional methods, such as geometric path planning and fuzzy logic controllers, in both adaptability and efficiency. These results affirm the potential of SAC as a scalable and versatile approach for advancing autonomous parking technology, while also highlighting avenues for future work, such as integrating hybrid methodologies and transitioning to real-world deployments to further validate and refine the system's practical applicability.